\pdfoutput=1

\documentclass[11pt]{article}

\usepackage[final]{acl}

\usepackage{times}
\usepackage{latexsym}

\usepackage[T1]{fontenc}

\usepackage[utf8]{inputenc}

\usepackage{microtype}

\usepackage{inconsolata}

\usepackage{graphicx}
\usepackage{tikz}
\usepackage{subcaption}

\usepackage{booktabs}
\usepackage{supertabular}
\usepackage{changepage}

\title{Culture is Not Trivia: Sociocultural Theory for Cultural NLP}

\author{Naitian Zhou,$^1$ David Bamman$^1$\and Isaac L. Bleaman$^2$ \\
  $^1$School of Information, University of California, Berkeley \\
  $^2$Department of Linguistics, University of California, Berkeley\\
  \texttt{\{naitian,dbamman,bleaman\}@berkeley.edu} 
  \\}

\begin{document}
\maketitle

\begin{abstract}
The field of cultural NLP has recently experienced rapid growth, driven by a pressing need to ensure that language technologies are effective and safe across a pluralistic user base. This work has largely progressed without a shared conception of culture, instead choosing to rely on a wide array of cultural proxies. However, this leads to a number of recurring limitations: coarse national boundaries fail to capture nuanced differences that lay within them, limited coverage restricts datasets to only a subset of usually highly-represented cultures, and a lack of dynamicity results in static cultural benchmarks that do not change as culture evolves. In this position paper, we argue that these methodological limitations are symptomatic of a theoretical gap. We draw on a well-developed theory of culture from sociocultural linguistics to fill this gap by 1) demonstrating in a case study how it can clarify methodological constraints and affordances, 2) offering theoretically-motivated paths forward to achieving cultural competence, and 3) arguing that localization is a more useful framing for the goals of much current work in cultural NLP.
\end{abstract}

\section{Introduction}

Language and culture are closely linked: language can be conceptualized
simultaneously as an artifact of culture as well as a process through which
culture is created \cite{ochsLinguisticResourcesSocializing2009}.
As language technologies become increasingly integrated into the everyday lives
of a diverse set of users, it is imperative that they are robust
to cultural differences between user bases
\cite{hershcovich-etal-2022-challenges}.

Cultural NLP, sometimes also known as cultural alignment, is a subfield within the NLP and ML communities that has experienced drastic growth in recent years to meet this challenge.
Work in cultural NLP usually involves building or evaluating systems that 1) have knowledge of cultural facts and 2) apply this knowledge appropriately in specific situations where cultural knowledge is relevant \cite{adilazuarda-etal-2024-towards,liuCulturallyAwareAdapted2024}.
This can include building new evaluation benchmarks or fine-tuning datasets that contain cultural knowledge of some kind, either manually \cite{lee-etal-2024-kornat,koto-etal-2024-indoculture} or automatically from a large corpus \cite{shi-etal-2024-culturebank,wang-etal-2024-craft}, or creating systems that generate culturally-relevant output \cite{khanuja-etal-2024-image}.

Most work relies on various proxies for defining both cultural boundaries and cultural objects.
Proxies of cultural boundaries commonly include nationality, religion, ethnicity, or other demographic features.
Proxies for cultural objects might include culture-specific knowlege of foods, values, or norms \cite{zhouDoesMapoTofu2024,sorensenValueKaleidoscopeEngaging2024,dwivediEtiCorCorpusAnalyzing2023}.
These works constitute an important step forward in understanding how to build fairer, more inclusive language technologies.
However, the disparate array of cultural proxies being evaluated is symptomatic of a theoretical gap: to achieve culturally-competent NLP systems, we must make progress towards a clearer, more unified conception of culture, and what it means for the systems we build to be responsive to that.

Fortunately, cultural NLP is not alone in the search for a useful notion of culture, and its theoretical challenges are not new.
Dissatisfaction with the coarseness of demographic cultural boundaries led to the second wave of sociolinguistics, which refocused efforts on identifying local cultural meaning within communities of practice \cite{eckertThreeWavesVariation2012}.
Larger questions, like the utility of the culture concept, have been debated in fields like sociocultural anthropology, where some researchers have abandoned culture altogether as being essentializing and othering \cite{vann2013culture}.
Indeed, epistemological and empirical tensions as they relate to the study of culture have been grappled with across such fields as anthropology, sociolinguistics, sociology, cultural studies, among many others.

\paragraph{Contributions.}

In this paper, we draw on theoretical developments in \textit{sociocultural linguistics} \cite{bucholtzIdentityInteractionSociocultural2005} --- itself a collection of several adjacent disciplines ---  to clarify the status of cultural knowledge in building culturally competent NLP systems.

We first review the goals of cultural NLP, and enumerate specific desiderata for culturally-aware language technologies that current works pursue (\S\ref{sec:goals}). Then, we highlight recurring difficulties in cultural NLP (\S\ref{sec:troubles}) by providing a survey of common self-stated limitations in existing papers. We introduce sociocultural linguistics as a field with a useful theoretical framework which we can apply to better understand culture as an object of study (\S\ref{sec:sociocultural_solution}), and provide a case study to illustrate how the theory of \textit{indexicality} can be applied to clarify the distinction between learning cultural knowledge and learning stereotypes (\S\ref{sec:case_study}). Finally, we discuss the broader implications granted by this understanding of culture, offering two main claims. First, we highlight existing methodological and theoretical gaps in achieving the ambitious goal of cultural competence, and provide theoretically-motivated suggestions for making progress on the task (\S\ref{sec:cult_competence}). Then, we argue that the goal in cultural NLP might be reasonably understood as localization instead of cultural competence or understanding, providing a more tractable and situated framing with which to build useful NLP systems (\S\ref{sec:localized_nlp}).

\section{The goals of cultural NLP}
\label{sec:goals}

\newcounter{circlecounter}
\setcounter{circlecounter}{1}

\newif\ifshowcircles
\showcirclestrue  %
\showcirclesfalse %

\newcommand{\numberedcircle}[1]{%
    \ifshowcircles
        \begin{tikzpicture}[baseline={([yshift=-.7ex]current bounding box.center)}]
            \node[
                circle,
                draw,
                inner sep=0.1em,
                text=black,
                font=\fontsize{0.7em}{0.7em}\bfseries\ttfamily\selectfont
            ] {#1};
        \end{tikzpicture}\nobreak \nobreak
    \fi
}

\newcommand{\autonumberedcircle}{%
    \ifshowcircles
        \numberedcircle{\thecirclecounter}%
        \stepcounter{circlecounter}%
    \fi
}

\begin{figure}
    \centering
\begin{subfigure}{0.48\linewidth}
\centering
\begin{tikzpicture}[>=stealth,scale=.9,baseline]
    \draw[gray!20] (0,0) grid (2.9,2.9);
    
    \draw[->] (-0,0) -- (3,0);
    \draw[->] (0,-0) -- (0,3);
    
    \node[rotate=90,above] at (0,1.5) {Adaptive};
    \node[below] at (1.5, 0) {Discerning};

    \draw[fill=olive!80, smooth cycle,draw=none,tension=0.8,opacity=0.3] 
      plot coordinates {(0.3, 2.4) (1.0,1.4) (2.3, 0.4) (0.9, 0.7) (0.4, 1.4)};

    \node at (0.7, 2) {\numberedcircle{1}};
    \node at (2, 0.6) {\numberedcircle{2}};
    \node at (2.6, 0.3) {\numberedcircle{3}};
    \node at (0.5, 1.2) {\numberedcircle{4}};
    \node at (0.3, 2.1) {\numberedcircle{5}};

\end{tikzpicture}
\end{subfigure}
\begin{subfigure}{0.48\linewidth}
\centering
\begin{tikzpicture}[>=stealth,scale=.9,baseline]
    \draw[gray!20] (0,0) grid (2.9,2.9);
    
    \draw[->] (-0,0) -- (3,0);
    \draw[->] (0,-0) -- (0,3);
    
    \node[rotate=90,above] at (0, 1.5) {Inclusive};
    \node[below] at (1.5, 0) {Nuanced};

    \draw[fill=olive!80, smooth cycle,draw=none,tension=0.8,opacity=0.3] 
      plot coordinates {(0.3, 2.4) (1.0,1.4) (2.3, 0.4) (0.9, 0.7) (0.4, 1.4)};

    \node at (0.4, 0.7) {\numberedcircle{1}};
    
    \node at (1.2, 2) {\numberedcircle{2}};  

    \node at (0.9, 2.2) {\numberedcircle{3}};

    \node at (0.3, 1.5) {\numberedcircle{4}};
    \node at (2.1, 0.4) {\numberedcircle{5}};    

\end{tikzpicture}
\end{subfigure}
\caption{Two separate spaces of desiderata; the \textbf{left} represents aspects of \textit{competence}, while the \textbf{right} represents aspects \textit{coverage}.
In each space, there often exists a trade-off between the axes, so that most cultural NLP work falls into the conceptual area that is shaded.}
\label{fig:desiderata}
\end{figure}
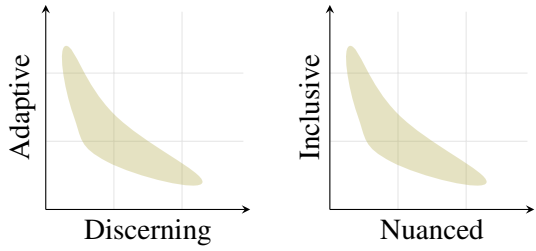

Though the field of cultural NLP does not necessarily agree on a definition of culture, there is general agreement on the goal:
to build culturally-\textit{competent} NLP systems \cite{bhatt-diaz-2024-extrinsic}.
Here, we break down this high-level goal into several more specific desiderata that are frequently mentioned in cultural NLP papers. We want our language technologies to be:

\paragraph{Adaptive.} 
A foundational premise of cultural NLP is that language technologies should be culturally \textit{sensitive}. In other words, culturally competent language technologies should be responsive to specific cultural contexts when designing their outputs. It would be insufficient for an NLP system to produce the same output for all cultural contexts; many works on bias in NLP have shown and problematized the tendency of language technologies to represent, exaggerate, and perpetuate a hegemonic set of values and structures \cite{voigt-etal-2018-rtgender,sheng-etal-2019-woman,benderDangersStochasticParrots2021}.

A wide body of work focuses on assessing whether NLP systems can generate different, appropriate outputs in response to different cultural contexts --- by answering value-oriented survey questions in a manner consistent with the target culture, for example \cite[\autonumberedcircle][]{cao-etal-2024-bridging,huang-yang-2023-culturally}. Some works that center adaptation as a value focus on \textit{extrinsic} evaluation: instead of probing whether language models \textit{know} specific cultural facts, they test whether NLP systems \textit{respond} in a way that demonstrates this knowledge \cite{bhatt-diaz-2024-extrinsic}.

\paragraph{Discerning.} At the same time, there is also a desire that NLP systems not perpetuate reductive stereotypes. Past work has demonstrated that users have differing expectations for cultural adaptation \cite{lucyOneSizeFitsAllExaminingExpectations2024}, and that cultural adaptation is not equally desired in all settings. For example, users may want technologies to understand their regional or ethnic dialects, but not generate them \cite{blaschke-etal-2024-dialect}.
This has motivated work on stereotype mitigation \cite[\autonumberedcircle][]{jha-etal-2023-seegull,maIntersectionalStereotypesLarge2023}, in which datasets of harmful stereotypes are collected in order to evaluate or engineer systems to avoid generating them.

\paragraph{Inclusive.} Cultural NLP values \textit{breadth}: language technologies should perform well across a large number of cultures. %
This value is represented by many works in the genre which build benchmarks for a large number of different cultures \cite[\autonumberedcircle][]{bhutani-etal-2024-seegull}; these papers usually use nationality and surveys as tractable ways of achieving large scale \cite[\autonumberedcircle][]{zhao-etal-2024-worldvaluesbench,ramezani-xu-2023-knowledge}. Some text mining methods for accumulating cultural knowledge also reflect the emphasis on large-scale, broad coverage \cite{fungMassivelyMultiCulturalKnowledge2024,nguyenExtractingCulturalCommonsense2023}.

\paragraph{Nuanced.} In addition to breadth, there is a desire for depth in the form of more granular and extensive cultural understanding. The value of nuance motivates works which build resources for specific languages or contexts \cite{koto-etal-2024-indoculture,son-etal-2025-kmmlu,li-etal-2024-cmmlu} and explore locally meaningful cultural categories \cite[\autonumberedcircle][]{devBuildingSocioculturallyInclusive2023}. These works might rely on local informants to provide cultural knowledge \cite{koto-etal-2024-indoculture}, trading breadth of coverage for richer cultural knowledge that large-scale survey-based methods cannot capture.%

\vspace{1em}\noindent
Though these four desiderata are not mutually exclusive, they roughly coalesce into two sets of trade-offs (visualized in fig.~\ref{fig:desiderata}).
The first two desiderata reflect two kinds of cultural \textit{competence}: the knowledge of how to respond differentially, and the knowledge of when it is appropriate to do so.
The second two desiderata reflect an orthogonal value of cultural \textit{coverage}: we want systems that cover many cultures, as well as systems that cover many aspects of each culture.

\section{Recurring troubles}
\label{sec:troubles}

Explicitly stating these desiderata can shed light on the motivations of current work, but they do not themselves offer any answers about what ``culture'' is. This becomes apparent when we look into the limitations sections of many cultural NLP papers, where
we find recurring themes that point to challenges posed by overly narrow definitions of culture. We survey the self-stated limitations of 57 papers from 2022-2024 which explicitly mention culture, as well as the cultural proxies they use.\footnote{The full list of papers can be found in the appendix.} This is not meant to be an exhaustive survey, but rather illustrative of the general state of the field.

The most commonly cited limitation was one of coverage (40\% of papers): the dataset or evaluation being presented was only collected with respect to a small subset of cultures. Partly, this can be explained by the proxies being used for setting cultural boundaries.
Of the papers we surveyed, 36\% of them used nationality as a demographic proxy \cite{adilazuarda-etal-2024-towards}. However, many papers problematize this choice in the limitations section as lacking \textit{nuance}, since nations are politically defined and not culturally homogeneous \cite{mendezBorderPoliticsContests2014}, and language labels usually reflect a hegemonic notion of a standard variety \cite{lippi-greenStandardLanguageMyth2011}.

Another common limitation was a lack of dynamicity (12\% of papers): culture is constantly constructed through social negotiation \cite{ochsLinguisticResourcesSocializing2009}, but benchmarks are largely static collections of examples or facts \cite{son-etal-2025-kmmlu,keleg-magdy-2023-dlama,jin-etal-2024-kobbq,li-etal-2024-cmmlu}. In most works, there is no granularity in the temporal dimension, failing to achieve an aspect of the desired \textit{nuance}. Uncertainty around the definition of culture also limits nuance and \textit{inclusivity} in cultural technologies, since most papers focus only on a small subset of culturally-relevant objects through proxies like food, etiquette, or values, without a framework to unify them.
Roughly 37\% of papers directly problematize their choice of a particular cultural proxy as being limited in its ability to represent culture as a whole, or too coarse to capture intragroup variation (28\% of papers mention this specifically).

Finally, the tension between \textit{adaptation} and \textit{discernment} results in uncertainty about how to address stereotypes in data. Some papers (14\%), which are largely intended for use in aligning models, view the potential of collecting stereotypes as cultural knowledge to be a limitation \cite{shi-etal-2024-culturebank}. Other papers explicitly collect stereotypes in order to build systems which can avoid generating them \cite{bhutani-etal-2024-seegull}.

Other limitations mentioned in various works include an overemphasis on English-language data and methods, the lack of extrinsic evaluation in favor of multiple-choice knowledge tests, the use of pretrained models to construct datasets, and various concerns with crowdsourced or human-annotated data, including the possibility that individual preferences are being construed as cultural ones.

Little progress has been made on rigorously addressing these limitations.
In many of these instances, there are questions in clear need of theoretical answers: how do we move past static, global categories when defining culture; how do we conceptualize culture in a way that respects its dynamic and constructed quality; how can we unify different facets of culture; how do we appropriately model and study stereotypes to build fairer systems?

\section{A sociocultural solution}
\label{sec:sociocultural_solution}
The notion of cultural competence is most commonly referenced in social and health services research, where culturally competent care has been encouraged as a way to reduce disparities in care quality and outcomes \cite{alizadehCulturalCompetenceDimensions2016}. Similar to cultural NLP, these works face operational challenges in identifying what cultural competence should look like \cite{kirmayerRethinkingCulturalCompetence2012}. Often, this literature draws on sociological and anthropological work to resolve these challenges. We will do the same here, focusing our attention on the fields of linguistic anthropology and sociolinguistics, which study language and culture in tandem.

For computational linguists, cultural competence might evoke the notion of linguistic competence, or Gumperz's more general idea of \textit{communicative competence} \cite{gumperzCommunicativeCompetence1997}: the ``knowledge of linguistic and related communicative conventions that speakers must have to initiate and sustain conversational involvement.'' Gumperz argues that communication must be understood not only in the context of linguistic systems of grammar, but within a semiotically rich social space. This idea has been widely accepted and refined in both linguistic anthropology and sociolinguistics,\footnote{Indeed, Gumperz was greatly influential in the establishment and progress of both these fields.} and we take \textit{social context} as a point of departure for understanding \textit{culture}.

\subsection{Text and context}
\label{sec:text_and_context}
Insofar as culture can be construed as a structured social phenomenon, it makes sense to understand it as the aspects of (extralinguistic) social context which make themselves interactionally relevant. In Gumperz's terms, this context contributes to a more general level of sensemaking in an interaction. \citet{edwardsCategoriesAreTalking1991} points out that even our linguistic categories are not subject only to cognitive processes, but also to social ones: the semantic category of ``bird'' might evoke an image of a robin or sparrow in a test-taking setting, but certainly indexes a different one at Thanksgiving dinner.\footnote{In the United States, turkey is often a centerpiece in the Thanksgiving meal.}

Addressing culture as social context shifts the ambiguity from one term to the other. The question becomes, what do we take to be social context? It is useful to look at the evolution of sociolinguistics as another quantitative discipline in which this question is at the fore. Early sociolinguistic studies focused frequently on sociological categories like socioeconomic class \cite{labovHypercorrectionLoweMiddle1985,guyLanguageSocialClass2011} or gender \cite{lakoffLanguageWomansPlace1973}, placing the speaker as a passive member of an externally-imposed category \cite{eckertThreeWavesVariation2012}. 
This led to concerns about the limits of coarse macrosociological categories, much like the critique of nationality in cultural NLP today.
In response, the second wave of sociolinguistic research incorporated ethnographic methods to better understand \textit{local} dynamics of language variation, relying on social networks and locally-relevant social categories. While second wave studies focused on local meaning, they still treated social categories as static, an essentializing assumption that equates identity with group affiliation. Third wave studies focus on \textit{identity} as a performance constructed from a diversity of semiotic resources including, but not limited to, language style. Social context, then, becomes the space within which identity is constructed and performed.

This evolution represents not only theoretical developments in response to empirical challenges in sociolinguistics, but also a steady convergence of ideas with other disciplines.
\citet{bucholtzIdentityInteractionSociocultural2005} provide a well-integrated framework for analyzing language and sociocultural identity in the form of \textit{sociocultural linguistics}, which synthesizes a convergent set of ideas from across disciplines to analyze language as well as other semiotic practices. Thus, sociocultural linguistics should not be thought of as a single theory of culture, but rather a concordant collection of theories that have seen relative convergence across fields that study language, culture, and society. 

We take this framework as the point of departure for the rest of this paper. We describe some foundational concepts (\S\ref{sec:primer}) and provide a case study for how they can clarify the objects and goals of cultural NLP (\S\ref{sec:case_study}). Then, we take a broader look at how sociocultural linguistic theory can inform computational work on cultural competence (\S\ref{sec:cult_competence}) and advance the goals of cultural NLP (\S\ref{sec:localized_nlp}).

\subsection{A primer on sociocultural linguistics}
\label{sec:primer}

\citet{bucholtzIdentityInteractionSociocultural2005} lay out five core principles of sociocultural linguistics. The terminology they use centers on the concept of \textit{identity}, defined as ``the social positioning of self and other.'' If we take the analogy of identity as a social position, we might consider culture to be the broader landscape within which identities are located. Through this lens, it becomes clear that individual identities not only \textit{reflect}, but also \textit{constitute} culture, giving shape to the cultural terrain. Thus, sociocultural linguistics provides a useful way of conceptualizing the cultural system more generally.\footnote{While \citet{bucholtzIdentityInteractionSociocultural2005} is nominally centered around identity and interaction, they in fact discuss culture more generally at various points (e.g., when introducing partialness).}

\paragraph{Emergence.} Identity emerges through interaction. This is the view that language does not \textit{come from} culture, but rather that culture is constituted \textit{through} linguistic (and other forms of) interaction. This draws on, among others, the ideas of identity performance \cite{butlerPerformativeActsGender1988} and audience design \cite{bellLanguageStyleAudience1984}. Emergence supports a more \textit{nuanced} representation of culture as one that is dynamic, and a more complex notion of \textit{adaptation} that supports the idea that even an individual can inhabit multiple cultural roles. This offers one solution to the challenge of defining cultural categories: it may make sense to instead induce cultural categories latent in the data.

\paragraph{Positionality.} Identity includes multiple levels of categories, including macro-level demographics, locally-specific distinctions, and contextually-specific stances and styles. This is a more \textit{inclusive} and \textit{nuanced} notion of culture that speaks to one of the core limitations of current work. National identity is only one level at which identity occurs, and the idea of positionality insists that we understand more granular categories of identity as well, including ones that are local to a specific community or even an interaction.

\paragraph{Indexicality.} Identity is constructed through an indexical process involving signs and their conceptual referents \cite{silversteinIndexicalOrderDialectics2003,eckertVariationIndexicalField2008}. Indexicality offers a \textit{mechanism} through which culture is constructed; it is the process of drawing links between linguistic (and other) forms and social meaning. These can play into cultural ideologies about language, construct stances local to specific interactions, consist of overt references to identity, and more; this provides a unified mechanism through which we can conceptualize culture. In section~\ref{sec:case_study}, we explore an example of how indexicality provides a useful theoretical account of stereotype in cultural NLP.

\paragraph{Relationality.} Identity takes on social meaning in relation to other identities. This provides a useful way of conceptualizing culture that aligns with findings in machine learning: contrastive learning of feature spaces often result in stronger representations than supervised learning among predefined categories. Similarity and difference are not the only relations available within this framework, which also includes authentication-denaturalization and authorization-illegitimization, among others. If the cultural space is structured through these relations, computational methods might benefit from considering how to encode them. 

\paragraph{Partialness.} Finally, any account of culture is necessarily incomplete, since it is itself situated contextually in relation to the subject it describes. Indeed, a person's identity at a given point in time may be partially deliberate, partially habitual (and subconscious), partially attributable to perception, partially conditioned by the interactional context, and partially subject to the ideologies that surround the interaction. That there is no single ground truth is a troubling statement for those who want to build robust, generalizable systems. However, it also provides a certain freedom from a positivist mirage. Instead, researchers and system designers are encouraged to think more critically about their position, and the assumptions encoded in the technology they build, with respect to the users whom these systems impact.

\section{Case study: culture from text}
\label{sec:case_study}

In \S\ref{sec:text_and_context}, we distinguish between language (the text) as the traditional object of study in linguistics and various ways of assessing culture as the surrounding social context. It has become essentially paradigmatic within NLP that we should expect to derive extratextual information (such as world models, for example) from training on text alone. It is reasonable, then, that there is a vein of cultural NLP work that mines for facts about specific identities from culturally-centered discourse on social media or the Internet \cite{rao-etal-2025-normad,fungMassivelyMultiCulturalKnowledge2024,nguyenExtractingCulturalCommonsense2023}.
In some cases, large language models are prompted to \textit{generate} specific culturally-relevant scenarios \cite{qiu-etal-2024-evaluating}. In many of these papers, authors note the dangerous potential for extracting biased or stereotyping information. How should we understand the epistemic status of these surfaced facts as cultural knowledge?
\citet[314]{labovSociolinguisticPatterns1973} defines stereotype as the linguistic forms which are subject to metapragmatic discussion: that is, the signs whose meanings are actively discussed.
In this section, we apply indexical theory to demonstrate that these works, in fact, can \textit{only} learn stereotypes.

Papers that mine for cultural facts aim to construct an indexical field from unstructured web text.
These systems take as input online documents or discussions about cultural differences (e.g., a guide to dining etiquette in India; \citealp{rao-etal-2025-normad}) and extract indexical mappings resembling this form:
\begin{quote}
    In \textbf{cultural group}, \textbf{belief} is widely accepted.
\end{quote}
This is effectively a mapping between the space of beliefs and the cultural groups that they index. Indexicality can occur at different levels of social awareness; a \textit{first-order} index evidences membership in a group. For example, the use of ``pop''  over ``soda'' might index membership in the population of Midwestern U.S. English speakers.\footnote{In \textbf{the Midwest}, it is widely accepted that \textbf{fizzy, sugary drinks are called ``pop.''}} However, higher-order indices occur as these associations themselves become embedded in cultural ideology. As such, the understanding that Midwesterners say ``pop'' is \textit{in and of itself} a piece of cultural knowledge \cite{eckertVariationIndexicalField2008}.\footnote{In \textbf{the U.S.}, it is widely accepted that \textbf{Midwestern U.S. English speakers use ``pop'' over ``soda.''}}

\subsection{Stereotypes all the way down}
\label{sec:high-order}

The implication of this idea is that works which study cultural discourse are \textit{primarily studying the stereotypes} embedded in the ideologies of the groups that generated this data, and only incidentally studying the cultures that are the objects of discourse. The status of these cultural facts as stereotypes does not depend on whether the group generating the discourse is the same as the group serving as the subject: the fact that these are cultural associations being \textit{discussed} rather than \textit{observed} classifies them as stereotypes. In fact, groups can, and often do, have both positive and negative stereotypes for themselves \cite{leavitt_frozen_2015,coffman_evidence_2014,sinclair_self-stereotyping_2006,pickett_motivated_2002}.

By applying the theory of indexical order, we gain clarity on the aspects of culture being studied. In the case of papers that mine cultural knowledge from cultural discourse, we find the subjects of study to be different from what we initially assumed. We are not learning about a diverse set of international cultures, but rather the world-view of the text authors, situated in a specific interactional context (perhaps posting about culture shock).

This is not just a matter of naming, nor a dismissal of the utility of these datasets. Instead, indexical theory clarifies the extent to which they are useful. It shows that these datasets exclude, by construction, cultural knowledge that is \textit{not} subject to metapragmatic discussion. It shows that higher-order indices can still be useful because their meanings are tied to the lower-order ones from which they arise. But it also illustrates complications that we must contend with: higher-order indices might persist even when lower-order ones are no longer as salient. ``Authentic'' Pittsburghers, for example, might be described as unpretentious, hospitable, sports-loving, etc. But this style originally indexed the immigrant-heritage, working-class history of the formerly industrial city, an identity that may not necessarily apply to its current residents, many of whom work in the health-care or higher-education sectors  \cite{johnstone100AuthenticPittsburgh2014}.

This is also not to say that first-order indices are inaccessible through computational methods. For example, computational sociolinguistic work successfully identifies first-order social meaning by using platform metadata like Twitter geolocation \cite{grieve_mapping_2019} or subreddit \cite{zhangCommunityIdentityUser2017,lucy-bamman-2021-characterizing} to measure sociolinguistic variation in linguistic features (often lexical or semantic) and the sociocultural meaning they index (like locale or community).

\subsection{Indexical values are contextual}
\label{sec:interacting}

It is also a mistake to assume that a given style from a given speaker always indexes the same thing, because the indexical value is also dependent on context, and interactionally interpreted.

\citet{chunMeaningMockingStylizations2007}, for example, provides an account of a ``foreign speaker'' language style as deployed by Asian American high schoolers.
She notes how this style can be employed both as accommodation to foreign speakers (e.g., a child speaking to her immigrant parents) and as mockery (e.g., between two peers at school). Sometimes, quotatively, an utterance can even fulfill both roles depending on the interactional frame through which it is interpreted.
The social meaning of an utterance is determined situationally within a specific interactional context.

\section{Paths forward}
\label{sec:paths}

Sociocultural linguistics paints a picture of culture as a complex, dynamic system through which sense-making occurs.
It is one that has proven useful in accounting for and describing how semiotic systems are constructed and deployed for social action in everyday interactions.

\subsection{Culturally competent NLP}
\label{sec:cult_competence}

But there exists a gap between this model of culture and our current computational methods for approaching culture. There is opportunity for NLP work to fill in these gaps.

Sociocultural linguistic theory tells us that culture is \textit{emergent}, and cultural NLP acknowledges that culture is a dynamic process, but currently our datasets are limited to static snapshots of cultural artifacts.
It may be fruitful to instead analyze discursive sequences in which cultural knowledge is suggested or contested.
When and how are norms enforced in interaction? How is cultural knowledge shared, and how is it taken up by the rest of the community? As an example, consider this interaction between two Latina high school students from \citet{mendoza-dentonHomegirls2008}:
\begin{adjustwidth}{-0.5cm}{}
\begin{quote}
\vspace{1em}  %
\begin{tabular}{p{0.25\linewidth}@{ }p{0.65\linewidth}}
    \textbf{Lupe:} & ¿Qué me ves? \\ & \textit{(What are you looking at?)} \\
    \textbf{Patricia:} & Tschhh, don’t EVEN talk to me in Spanish, ‘cause your Spanish ain’t all that.
\end{tabular}
\vspace{1em}  %
\end{quote}
\end{adjustwidth}
Through contextual information like the participants' posture, make-up, and social networks (including the fact that they are rival gang members), we can understand the setting of this interaction: Patricia has interpreted Lupe's question to be a claim to authenticity.
But through the interaction itself we can see the cultural process in action: as \citet{mendoza-dentonHomegirls2008} notes, Lupe asserts her Mexican-ness symbolically through her use of Spanish. Through both her assertion and Patricia's contestation, the social importance of Spanish is \textit{reinforced} as indexing their Mexican identities.
Analogous computational work might study comment threads for these kinds of interactions, and additionally incorporate contextual mechanisms like flairs or voting that users can employ to express affiliation or pass judgment on platforms like Reddit \cite{gaudetteUpvotingExtremismCollective2021}.

Sociocultural theory tells us that culture is \textit{positional}, operating at multiple levels of identity and often composed of features from many different styles, but current methods impose coarse, usually unidimensional, boundaries like nationality on cultural categories.
Relationality and indexicality offer mechanisms through which cultural sensemaking occurs --- how can we better model positionality as a contextually legible field of identities by identifying instances of cultural categories being constructed in relation to other categories, or identities being assembled by combining different indexical signs?
\citet{castelleSapirsThoughtGroovesWhorfs} suggests that modern language models can be usefully conceptualized more generally as effective learners of semiotic systems; how might we build systems that learn representation spaces for other kinds of meaning beyond semantics, like social or discursive meaning?

Indexicality also motivates the need for datasets that are contextually rich: culture is the combination and construction of different semiotic resources that make reference to social meaning, yet our methods are deployed on datasets that largely consist of decontextualized text.
Data that contains social context in other forms (e.g., metadata or other kinds of world state) could be one way of addressing this limitation \cite{nguyenCollaborativeGrowthWhen2025}. For example, the STAC corpus \cite{asher-etal-2016-discourse} consists of dialogue situated in a game scenario, and includes information about the game state and actions. This places linguistic interaction within a broader context; future works might extend this paradigm to other, more socially relevant metadata.

Indexical fields also exist beyond text, reaching into other modalities in the form of gesture, prosody, and even extralinguistic semiotic systems like fashion \cite{chunMeaningMockingStylizations2007} and make-up \cite{mendoza-dentonHomegirls2008}.
Not only is it important to represent non-text modalities to capture culture, but combining modalities can also be a promising direction to learning social meaning \cite{zhou-etal-2024-social}.

There is also much theoretical work at hand to account for how a software system might differ from a human in how it is taken up as an interlocutor in interaction. Creating systems that perfectly replicate human behavior is neither desirable nor felicitous. Consider this podcast transcript introducing the findings of a scientific paper:

\begin{adjustwidth}{-0.5cm}{-0.5cm}  %
\begin{quote}
\vspace{1em}  %

\begin{tabular}{p{0.2\linewidth}@{ }p{0.75\linewidth}}
    \textbf{Host A:} & Think about those old Hollywood films, the ones your grandma might watch. \\
    \textbf{Host A:} & Do those performances feel different than what you might see in movies today? \\
    \textbf{Host B:} & Hm, yeah I guess they do. It's, like, more dramatic. The emotions are way more, out there? \\
\end{tabular}
\end{quote}
\vspace{1em}  %

\end{adjustwidth}
This serves the discursive purpose of simultaneously motivating a finding and establishing rapport with the listener by drawing on the presenter's personal experience. However, if this same script is generated with an LLM,\footnote{As, indeed, it was, by NotebookLM \cite{googleNotebookLM2024}.} the social action becomes infelicitous. The LLM has no grandmother, whose past movie-going experiences are being imagined and described. Instead, the audience must reinterpet this sequence as a post-hoc rationalization of the source material that is about to be presented, failing to motivate the finding or establish rapport.
Not all semiotic resources available to humans are available to the language technologies we build.

\subsection{Localized NLP}
\label{sec:localized_nlp}

All told, we are far from building culturally competent systems, given these clear and pressing theoretical and methodological gaps.
But cultural NLP also faces more immediate goals, which are perhaps more central to the field as it is currently configured. We want to create, e.g., web agents that will not make food purchases that violate religious dietary laws \cite{qiu-etal-2025-evaluating} or image generation models that show the local currency when displaying money \cite{khanuja-etal-2024-image}. These goals are mediated by constraints on resources: user studies and qualitative interviews might provide richer cultural data, but are more expensive to scale. Do we need to achieve cultural competence in the general sense for these more immediate applications? And how do we navigate trade-offs between desiderata?

Many would argue that machine translation systems have not yet achieved linguistic competence (and this is perhaps an easier case to make in the multilingual setting). Yet, individual software applications have been internationalized long before MT achieved even its most recent success. When building systems that accommodate more users, a more useful, immediate framing might be one of \textit{localization} rather than cultural competence. Understanding the task at hand as building localized NLP applications helps us locate ourselves in the space of desiderata (fig.~\ref{fig:desiderata}).

Localization is tractable because it forces us to specify the application domain, constraining the relevant depth of knowledge. Localized translations are generated only for the necessary text within an application; culturally localized systems can focus on the domain-specific \textit{nuances} of cultural knowledge. Sociocultural approaches to culture are contextual and situated, and localization forces us to evaluate cultural performance in a situated application setting.

Localization also forces us to enumerate our audience, constraining and making explicit the \textit{coverage} of our systems.
Localized translations are not provided for an arbitrary, unconstrained set of languages or an arbitrary set of text.
Furthermore, a website that offers its interface in, e.g., ``Spanish'' rarely allows users to choose a specific regional dialect, even though different varieties of Spanish often show lexical and syntactic variation.
This is a pragmatic choice, but also an ideological one about which language varieties to support, and it is better that the ideological choices be made explicitly and transparently.
In the cultural setting, this can also make the choice of cultural boundaries less arbitrary. If the goal is to build a culturally localized healthcare chatbot, for example, differing levels of medical literacy may be a more salient cultural boundary with more actionable interventions than something like nationality.

Finally, localization forces us to consider the NLP system as an interlocutor in the human-computer interaction. While many existing cultural knowledge benchmarks probe large language models removed from the specific context of how they will be used, approaching the task as localization forces us to define the expected \textit{behavior} within a given application context. Developers of a recipe application might improve user experience by offering culture-specific ingredient substitutions \cite{she-etal-2024-mapo}, but a healthcare application might benefit from adopting a stance of cultural \textit{humility} instead of potentially stereotyping or stigmatizing \textit{adaptation} \cite{lekasRethinkingCulturalCompetence2020}. Defining the bounds of expected cultural performance specifies where we want a particular application to lie in the \textit{discerning / adaptive} space.

Thus, localization allows us to focus on particulars that may be more tractably implemented and evaluated in real-world systems today. Principles from user-centered \cite{lowdermilkUserCenteredDesignDevelopers2013} and participatory design \cite{caselliGuidingPrinciplesParticipatory2021}, particularly as applied to machine learning contexts \cite{ayobiComputationalNotebooksCoDesign2023,tanSeatTableEnough2024}, provide footholds for NLP practitioners who wish to identify the cultural features which are relevant to a given application context.

\section{Conclusion}

It is important to build language technologies that are responsive to cultural values. However, the current field of cultural NLP has not found agreement on what it means to model culture, settling instead for a wide array of cultural proxies for both categories of identity and categories of indexical features. In this paper, we deconstruct the goals of cultural NLP and highlight how recurring discomforts in current work are illustrative of a lack of theoretical alignment. We propose drawing on convergent theoretical insights from a variety of social-scientific disciplines which have centered the study of culture in the context of language and other semiotic systems.

When studying such a complex, multifaceted, and dynamic object as culture, it is equally challenging and imperative that the object of study be well-defined. We demonstrate how sociocultural linguistics provides a useful theoretical framework that treats culture as an enacted process, not a static artifact. We explore the implications of this: we show that learning cultural facts through metapragmatic discourse is limited to learning about stereotypes; we make the case that building culturally competent computational systems requires a dynamic model of culture as a process, not a collection of trivia, and that sociocultural linguistics provides a powerful model of this process, but methodological and theoretical gaps still loom large; finally, we argue that many current works in cultural NLP can be usefully reframed as localization, which encourages situated, participatory design and evaluation of systems.

The growth of cultural NLP reflects the more general state of natural language processing. Until recently, NLP has been largely preoccupied with learning the semantic meaning of textual symbols.
But other fields of linguistics have long established that the world around us cannot be extricated from the words we produce and interpret.
Gumperz argued fifty years ago that communicative competence reaches beyond grammatical knowledge.
As computational methods have become more powerful in representing textual semantic meaning, it becomes both tractable and necessary to consider other kinds of meaning, like sociocultural meaning, as equally important objects of study.

\section{Limitations}

Though we highlight certain representative works in making the case for a theory-led approach to cultural NLP, this paper is not meant to be a survey of the field. \citet{adilazuarda-etal-2024-towards} and \citet{liuCulturallyAwareAdapted2024} offer good overviews of recent work.

We also introduce the specific theoretical framework of sociocultural linguistics. Though other theories of culture certainly exist, we focus on sociocultural linguistics for its linguistically-oriented approach to culture (see \S\ref{sec:sociocultural_solution} for detailed discussion about this).

\section*{Acknowledgments}

This work was supported by funding from the National Science Foundation (Graduate Research Fellowship DGE-2146752).

We thank Laura Sterponi, Kent Chang, Dan Hickey, Nikita Mehandru, Lucy Li, and many others for their thoughtful discussion and feedback.

\bibliography{anthology,custom}

\newpage

\appendix

\section{List of surveyed works}

\tablehead{
  \toprule
  Paper & Limitations \\
  \midrule
}
\tabletail{
  \midrule
  \multicolumn{2}{r}{\small\textit{Table continues}} \\
}
\tablelasttail{
  \bottomrule
  \multicolumn{2}{r}{\small\textit{Table complete}} \\
}

\renewcommand{\arraystretch}{1.2}  %

\small
\begin{supertabular}{p{0.6\linewidth}p{0.3\linewidth}}

Good Night at 4 pm?! Time Expressions in Different Cultures \cite{shwartz-2022-good} & proxy, individual, multilingual \\
Probing Pre-Trained Language Models for Cross-Cultural Differences in Values \cite{arora-etal-2023-probing} & proxy, survey \\
GD-COMET: A Geo-Diverse Commonsense Inference Model \cite{bhatia-shwartz-2023-gd} & intrinsic, stereotype \\
Assessing Cross-Cultural Alignment between ChatGPT and Human Societies: An Empirical Study \cite{cao-etal-2023-assessing} & language as culture, proxy \\
Sociocultural Norm Similarities and Differences via Situational Alignment and Explainable Textual Entailment \cite{ch-wang-etal-2023-sociocultural} & coarseness, individual, coverage \\
Toward Cultural Bias Evaluation Datasets: The Case of Bengali Gender, Religious, and National Identity \cite{das-etal-2023-toward} & coverage \\
Building Socio-culturally Inclusive Stereotype Resources with Community Engagement \cite{devBuildingSocioculturallyInclusive2023} & context, coverage, multilingual \\
NORMSAGE: Multi-Lingual Multi-Cultural Norm Discovery from Conversations On-the-Fly \cite{fung-etal-2023-normsage} & stereotype, culture is dynamic \\
Multilingual Language Models are not Multicultural: A Case Study in Emotion \cite{havaldar-etal-2023-multilingual} & coverage, coarseness \\
Culturally Aware Natural Language Inference \cite{huang-yang-2023-culturally} & coarseness, stereotype \\
SeeGULL: A Stereotype Benchmark with Broad Geo-Cultural Coverage Leveraging Generative Models \cite{jha-etal-2023-seegull} & context, coarseness, subjectivity \\
Multi-lingual and Multi-cultural Figurative Language Understanding \cite{kabra-etal-2023-multi} & coarseness, unmarked culture \\
DLAMA: A Framework for Curating Culturally Diverse Facts for Probing the Knowledge of Pretrained Language Models \cite{keleg-magdy-2023-dlama} & proxy, coarseness, crowdsourced \\
Evaluating the Diversity, Equity, and Inclusion of NLP Technology: A Case Study for Indian Languages \cite{khanuja-etal-2023-evaluating} & proxy, language as culture, coarseness \\
NormMark: A Weakly Supervised Markov Model for Socio-cultural Norm Discovery \cite{moghimifar-etal-2023-normmark} & coverage \\
FORK: A Bite-Sized Test Set for Probing Culinary Cultural Biases in Commonsense Reasoning Models \cite{palta-rudinger-2023-fork} & proxy, coverage, annotator, reductive \\
Knowledge of cultural moral norms in large language models \cite{ramezani-xu-2023-knowledge} & coverage, culture is dynamic \\
NLPositionality: Characterizing Design Biases of Datasets and Models \cite{santy-etal-2023-nlpositionality} & proxy \\
Geographical Erasure in Language Generation \cite{schwobel-etal-2023-geographical} & multilingual \\
Modeling Cross-Cultural Pragmatic Inference with Codenames Duet \cite{shaikh-etal-2023-modeling} & stereotype \\
Cross-Cultural Analysis of Human Values, Morals, and Biases in Folk Tales \cite{wu-etal-2023-cross} & coverage, multilingual \\
Cross-Cultural Transfer Learning for Chinese Offensive Language Detection \cite{zhou-etal-2023-cross} & multilingual, coarseness \\
Cultural Compass: Predicting Transfer Learning Success in Offensive Language Detection with Cultural Features \cite{zhou-etal-2023-cultural} & proxy, coverage, coarseness \\
NormBank: A Knowledge Bank of Situational Social Norms \cite{ziems-etal-2023-normbank} & coverage, stereotype \\
Multi-VALUE: A Framework for Cross-Dialectal English NLP \cite{ziems-etal-2023-multi} & proxy \\
Investigating Cultural Alignment of Large Language Models \cite{alkhamissi-etal-2024-investigating} & coverage, proxy, reductive, coarseness \\
Extrinsic Evaluation of Cultural Competence in Large Language Models \cite{bhatt-diaz-2024-extrinsic} & proxy, multilingual \\
Bridging Cultural Nuances in Dialogue Agents through Cultural Value Surveys \cite{cao-etal-2024-bridging} & language as culture, pluralism \\
Cultural Adaptation of Recipes \cite{cao-etal-2024-cultural} & proxy, coarseness, coverage \\
The Echoes of Multilinguality: Tracing Cultural Value Shifts during LM Fine-tuning \cite{choenni-etal-2024-echoes} & language as culture, proxy, coarseness \\
Massively Multi-Cultural Knowledge Acquisition \& LM Benchmarking \cite{fungMassivelyMultiCulturalKnowledge2024} & LLM-derived \\
ViSAGe: A Global-Scale Analysis of Visual Stereotypes in Text-to-Image Generation \cite{jha-etal-2024-visage} & subjectivity, annotator, reductive, proxy, discerning \\
KoBBQ: Korean Bias Benchmark for Question Answering \cite{jin-etal-2024-kobbq} & subjectivity, proxy \\
CLIcK: A Benchmark Dataset of Cultural and Linguistic Intelligence in Korean \cite{kim-etal-2024-click} & proxy, coarseness, coverage, stereotype, discerning \\
The PRISM Alignment Dataset: What Participatory, Representative and Individualised Human Feedback Reveals About the Subjective and Multicultural Alignment of Large Language Models \cite{kirkPRISMAlignmentDataset2024} & discerning, preference v morality, exploited annotators \\
IndoCulture: Exploring Geographically-Influenced Cultural Commonsense Reasoning Across Eleven Indonesian Provinces \cite{koto-etal-2024-indoculture} & culture is dynamic, coverage \\
KorNAT: LLM Alignment Benchmark for Korean Social Values and Common Knowledge \cite{lee-etal-2024-kornat} & culture is dynamic, unmarked culture, annotator, intrinsic \\
CultureLLM: Incorporating Cultural Differences into Large Language Models \cite{liCultureLLMIncorporatingCultural2024} & intrinsic \\
CULTURE-GEN: Revealing Global Cultural Perception in Language Models through Natural Language Prompting \cite{liCULTUREGENRevealingGlobal2024} & intrinsic, coarseness \\
Are Multilingual LLMs Culturally-Diverse Reasoners? An Investigation into Multicultural Proverbs and Sayings \cite{liu-etal-2024-multilingual} & proxy, coverage \\
Cultural Alignment in Large Language Models: An Explanatory Analysis Based on Hofstede's Cultural Dimensions \cite{masoud-etal-2025-cultural} & culture is dynamic, multilingual, proxy \\
Having Beer after Prayer? Measuring Cultural Bias in Large Language Models \cite{naous-etal-2024-beer} & coarseness, coverage \\
Cultural Commonsense Knowledge for Intercultural Dialogues \cite{nguyenCulturalCommonsenseKnowledge2024} & crowdsourced, LLM-derived, stereotype \\
Evaluating Cultural and Social Awareness of LLM Web Agents \cite{qiu-etal-2025-evaluating} & multilingual, coverage, context \\
NormAd: A Framework for Measuring the Cultural Adaptability of Large Language Models \cite{rao-etal-2025-normad} & proxy, coarseness, culture is dynamic, multilingual, coverage \\
DOSA: A Dataset of Social Artifacts from Different Indian Geographical Subcultures \cite{seth-etal-2024-dosa} & coverage, intersectionality, multilingual \\
CultureBank: An Online Community-Driven Knowledge Base Towards Culturally Aware Language Technologies \cite{shi-etal-2024-culturebank} & multilingual, unmarked culture, stereotype \\
KMMLU: Measuring Massive Multitask Language Understanding in Korean \cite{son-etal-2025-kmmlu} & copyright, intrinsic \\
Value Kaleidoscope: Engaging AI with Pluralistic Human Values, Rights, and Duties \cite{sorensenValueKaleidoscopeEngaging2024} & LLM-derived \\
Navigating Cultural Chasms: Exploring and Unlocking the Cultural POV of Text-To-Image Models \cite{venturaNavigatingCulturalChasms2024} & coverage, based on existing model \\
CRAFT: Extracting and Tuning Cultural Instructions from the Wild \cite{wang-etal-2024-craft} & multilingual \\
SeaEval for Multilingual Foundation Models: From Cross-Lingual Alignment to Cultural Reasoning \cite{wang-etal-2024-seaeval} & coverage, intrinsic \\
Not All Countries Celebrate Thanksgiving: On the Cultural Dominance in Large Language Models \cite{wang-etal-2024-countries} & survey, coverage \\
Benchmarking Machine Translation with Cultural Awareness \cite{yao-etal-2024-benchmarking} & proxy \\
RENOVI: A Benchmark Towards Remediating Norm Violations in Socio-Cultural Conversations \cite{zhan-etal-2024-renovi} & multilingual \\
WorldValuesBench: A Large-Scale Benchmark Dataset for Multi-Cultural Value Awareness of Language Models \cite{zhao-etal-2024-worldvaluesbench} & culture is dynamic, proxy, individual \\
Does Mapo Tofu Contain Coffee? Probing LLMs for Food-related Cultural Knowledge \cite{zhouDoesMapoTofu2024} & coverage, multilingual, crowdsourced \\

\end{supertabular}

\section{Category key}

\tablehead{
  \toprule
  Category & Description \\
  \midrule
}
\tabletail{
  \midrule
  \multicolumn{2}{r}{\small\textit{Table continues}} \\
}
\tablelasttail{
  \bottomrule
  \multicolumn{2}{r}{\small\textit{Table complete}} \\
}

\begin{supertabular}{p{0.3\linewidth}p{0.6\linewidth}@{}}
\raggedright coverage & Makes note of insufficient or limited coverage. \\
\raggedright proxy & Notes the limitations of the chosen proxy. \\
\raggedright coarseness & Mentions that cultural boundaries are too coarse. \\
\raggedright multilingual & Mentions that only one (or a limited number) of languages are studied. \\
\raggedright stereotype & Mentions concerns about learning or perpetuating stereotypes. \\
\raggedright culture is dynamic & Mentions limitations in capturing the dynamicity of culture. \\
\raggedright intrinsic & Mentions the limitations of only performing intrinsic evaluation. \\
\raggedright language as culture & Mentions the limitations of treating language as cultural boundaries. \\
\raggedright crowdsourced & Mentions concerns about crowdsourced data. \\
\raggedright discerning & Mentions that not all cultural attributes should be treated the same way. \\
\raggedright LLM-derived & Mentions concerns that some part of the dataset was generated by LLMs. \\
\raggedright reductive & Mentions concerns that culture is reduced to the proxies chosen. \\
\raggedright annotator & Mentions concerns about human-annotated data. \\
\raggedright unmarked culture & Mentions concerns with ``universal'' cultural attributes or values. \\
\raggedright individual & Mentions that individual values in the data may not be reflective of cultural ones. \\
\raggedright subjectivity & Mentions that cultural annotations are subjective. \\
\raggedright context & Notes that the system or datasets lacks social or situational context. \\
\raggedright survey & Notes the limitations of relying on survey data for cultural knowledge. \\
\raggedright pluralism & Mentions concerns with choosing which value to align to in pluralistic situations. \\
\raggedright preference v morality & Notes that aligning to preferences may not be desirable behavior. \\
\raggedright exploited annotators & Notes that annotators providing preference data do not generally share in the benefits. \\
\raggedright intersectionality & Mentions limitations in covering intersectional identities. \\
\raggedright copyright & Notes that some data points were removed due to copyright. \\
\raggedright based on existing model & Notes that automated evaluation is based on an existing model. \\
\end{supertabular}

\end{document}